\documentclass[10pt,twocolumn,letterpaper]{article}

\usepackage{iccv}
\usepackage{times}
\usepackage{epsfig}
\usepackage{graphicx}
\usepackage{amsmath}
\usepackage{amssymb}

\usepackage{ifpdf}
\usepackage{graphicx} 
\usepackage{diagbox}
\usepackage{multirow}
\usepackage{color}
\usepackage{subfigure}
\usepackage{enumitem}
\usepackage{setspace}
\usepackage{booktabs}
\usepackage{url}
\setitemize{itemsep=0pt,partopsep=0pt,parsep=\parskip,topsep=3pt}

\usepackage[breaklinks=true,bookmarks=false]{hyperref}

\iccvfinalcopy 

\def\iccvPaperID{520} 

\ificcvfinal\fi

\begin{document}

\title{Weakly Aligned Cross-Modal Learning for Multispectral Pedestrian Detection}

\author{Lu Zhang$^{\rm{1,3}}$, Xiangyu Zhu$^{\rm{2,3}}$, Xiangyu Chen$^{\rm{5}}$, Xu Yang$^{\rm{1,3}}$, Zhen Lei$^{\rm{2,3}}$, Zhiyong Liu$^{\rm{1,3,4}}$\thanks{Corresponding author}\\
 $^{\rm 1}$ SKL-MCCS, Institute of Automation, Chinese Academy of Sciences\\
 $^{\rm 2}$ CBSR \& NLPR, Institute of Automation, Chinese Academy of Sciences\\
 $^{\rm 3}$ University of Chinese Academy of Sciences~
 $^{\rm 4}$ CEBSIT, Chinese Academy of Sciences\\
  $^{\rm 5}$ Renmin University of China\\
{\tt\small
\{zhanglu2016,xu.yang,zhiyong.liu\}@ia.ac.cn, \{xiangyu.zhu,zlei\}@nlpr.ia.ac.cn} 
}

\maketitle
\ificcvfinal\thispagestyle{empty}\fi

\begin{abstract}
Multispectral pedestrian detection has shown great advantages under poor illumination conditions, since the thermal modality provides complementary information for the color image. However, real multispectral data suffers from the position shift problem, \ie the color-thermal image pairs are not strictly aligned, making one object has different positions in different modalities. In deep learning based methods, this problem makes it difficult to fuse the feature maps from both modalities and puzzles the CNN training. In this paper, we propose a novel Aligned Region CNN (AR-CNN) to handle the weakly aligned multispectral data in an end-to-end way. Firstly, we design a Region Feature Alignment (RFA) module to capture the position shift and adaptively align the region features of the two modalities. Secondly, we present a new multimodal fusion method, which performs feature re-weighting to select more reliable features and suppress the useless ones. Besides, we propose a novel RoI jitter strategy to improve the robustness to unexpected shift patterns of different devices and system settings. Finally, since our method depends on a new kind of labelling: bounding boxes that match each modality, we manually relabel the KAIST dataset by locating bounding boxes in both modalities and building their relationships, providing a new KAIST-Paired Annotation. Extensive experimental validations on existing datasets are performed, demonstrating the effectiveness and robustness of the proposed method. Code and data are available at: \url{https://github.com/luzhang16/AR-CNN}.
\end{abstract}

\section{Introduction}

Pedestrian detection is an important research topic in computer vision field with various applications, such as video surveillance, autonomous driving, and robotics. Although great progress has been made by the deep learning based methods (\eg \cite{sermanet2013pedestrian, yang2015convolutional, zhang2016faster, li2018scale}), detecting the pedestrian in adverse illumination conditions, occlusions and clutter background is still a challenging problem. Recently, many works in robot vision \cite{burian2014robot, wang2019densefusion}, facial expression recognition \cite{corneanu2016survey}, material classification \cite{saponaro2015material}, and object detection \cite{song2016deep, deng2017amodal, gupta2014learning, xu2018pointfusion} show that adopting a novel modality can improve the performance and offer competitive advantages over single sensor systems. Among the sensors, thermal camera is widely used in face recognition \cite{buddharaju2007physiology, socolinsky2003face, kong2007multiscale}, human tracking \cite{leykin2007thermal, torabi2012iterative} and action recognition \cite{zhu2013study, gao2016infar} for its biometric robustness. Motivated by this, multispectral pedestrian detection \cite{hwang2015multispectral, xu2017learning, gonzalez2016pedestrian, Park2018Unified} has attracted massive attention and provides new opportunities for around-the-clock applications, mainly due to its superiority of complementary nature between color and thermal modalities.

\begin{figure}[!t]
\centering
\includegraphics[width=3.2in]{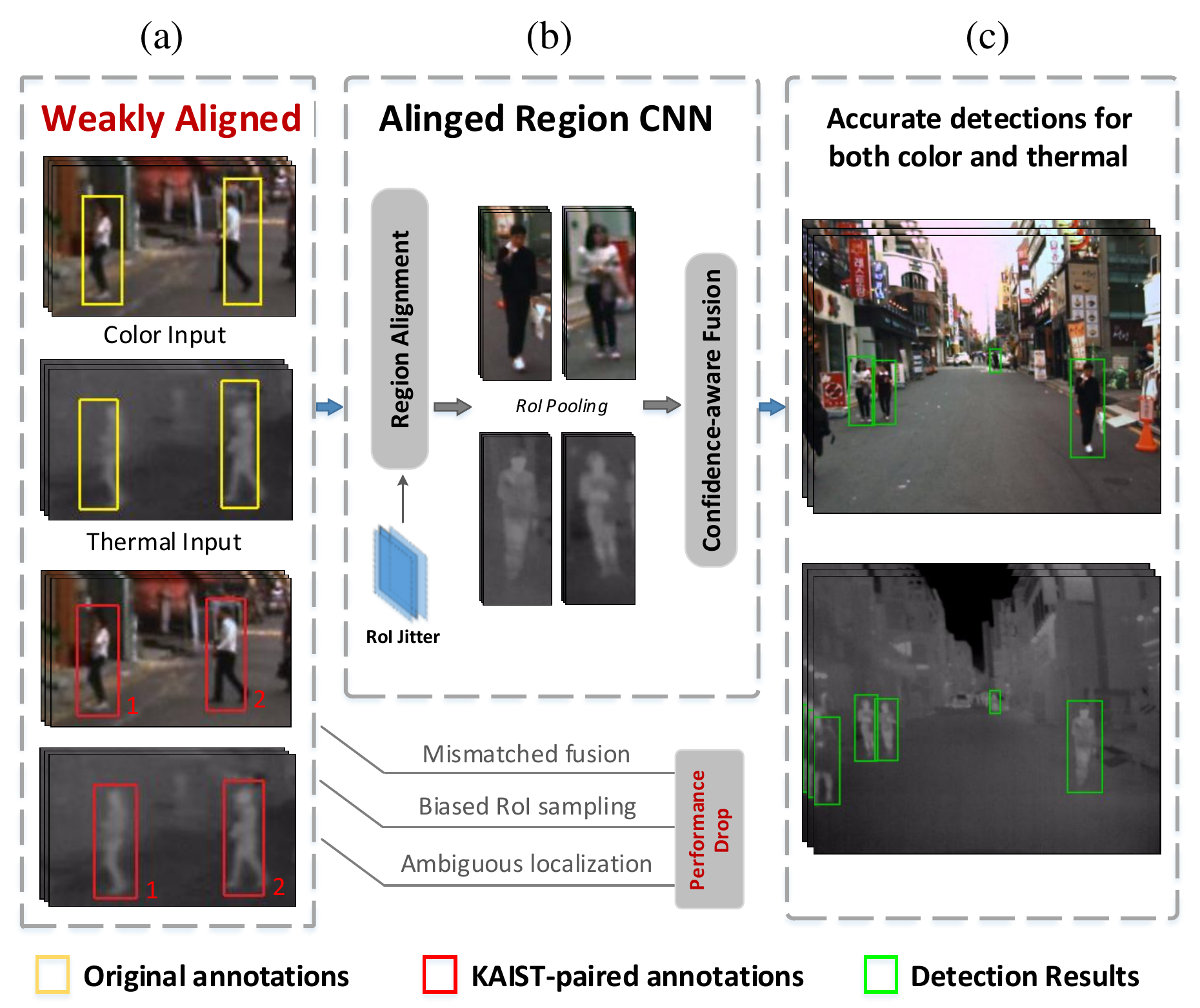}
\subfigure{\label{figure-overview-a}}
\caption{Overview of our framework. (a) The color-thermal pair and its annotations, the yellow boxes denote original KAIST annotations, which have position shift between two modalities; the red boxes are the proposed KAIST-Paired annotations, which have independent labelling to match each modality. (b) Illustration of the proposed AR-CNN model. (c) Detection results of both modalities under the position shift problem.}
\label{figure-overview}
\end{figure}

\textbf{Challenges}~~A common assumption of multispectral pedestrian detection is that the color-thermal image pairs are geometrically aligned \cite{hwang2015multispectral, xu2017learning,liu2016multispectral,konig2017fully,li2019illumination,guan2019fusion,zhang2019cross}.  
However, the modalities are just weakly aligned in practice, which means there is the position shift between modalities, making one object has different positions on different modalities, see Figure \ref{figure-overview-a}. This position shift problem can be caused by physical properties of different sensors (\eg parallax, mismatched resolutions and field-of-views), imperfection of alignment algorithms, external disturbance, and hardware aging. Moreover, the calibration for color-thermal cameras are tortuous, generally require particular hardware as well as special heated calibration board \cite{kim2015geometrical, hwang2015low, hwang2015multispectral, treible2017cats, choi2018kaist}. 

The position shift problem degrades the pedestrian detector in two aspects.  
First, features from different modalities are mismatched in the corresponding positions, which puzzles the inference. Second, it is difficult to cover the objects in both modalities with a single bounding box. In existing datasets \cite{hwang2015multispectral, gonzalez2016pedestrian}, the bounding box is either labelled on single modality (color or thermal) or a big bounding box is labelled to cover the objects on both modalities. This label bias will give bad supervision singals and degrade the detector especially for CNN based methods \cite{ren2015faster, liu2016ssd}, where the intersection over union (IoU) overlap is used for foreground/background classification. Therefore, how to robustly localize each individual person on weakly aligned modalities remains to be a critical issue for multispectral pedestrian detectors.

\textbf{Our Contributions}~~(1). To the best of our knowledge, this is the first work that tackles the position shift problem in multispectral pedestrian detection. In this paper, we analyse the impacts of the position shift problem and propose a novel detection framework to merge the information from both modalities. Specifically, a novel RFA module is presented to shift and align the feature maps from two modalities. Meanwhile, the RoI jitter training strategy is adopted to randomly jitter the RoIs of the sensed modality, improving the robustness to the patterns of position shift. Furthermore, a new confidence-aware fusion method is presented to effectively merge the modality, which adaptively performs feature re-weighting to select more reliable features and depress the confusing ones. Figure \ref{figure-overview} depicts an overview of the proposed approach.

(2). To realize our method, we manually relabel the KAIST dataset and provide a novel KAIST-Paired annotation. We first filter the image pairs with original annotations and obtain $20,025$ valid frames. Then $59,812$ pedestrians are carefully annotated by locating the bounding boxes in both modalities and building their relationships.

(3). The experimental results show that the proposed approach reduces the degradation from position shift problem and makes full use of both modalities, achieving the state-of-the-art performance on the challenging KAIST and CVC-14 dataset.

\section{Related Work}
\textbf{Multispectral Pedestrian Detection} 
As an essential step for various applications, pedestrian detection has attracted massive attention from the computer vision community. Over the years, extensive features and algorithms have been proposed, including both traditional detectors \cite{dollar2009integral, dollar2014fast, nam2014local, zhang2015filtered} and the lately dominated CNN-based detectors \cite{mao2017can, hosang2015taking, brazil2017illuminating, wang2018repulsion, zhang2018occlusion}. Recently, multispectral data have shown great advantages, especially for the all-day vision  \cite{kim2018multispectral, choi2016thermal, choi2018kaist}.
Hence the release of large-scale multispectral pedestrian benchmarks \cite{hwang2015multispectral, gonzalez2016pedestrian} is encouraging the research community to advance the state-of-the-art by efficiently exploiting multispectral input data. Hwang \etal \cite{hwang2015multispectral} propose an extended ACF method, leveraging aligned color-thermal image pairs for around-the-clock pedestrian detection.  With the recent development of deep learning, the CNN-based methods \cite{xu2017learning, wagner2016multispectral, choi2016multi, zhang2019cross, guan2018exploiting, li2019illumination} significantly improve the multispectral pedestrian detection performances. Liu \etal\cite{liu2016multispectral} adopt the Faster R-CNN architecture and analyze different fusion stages within the CNN. K{\"o}nig \etal\cite{konig2017fully} adapt the Region Proposal Network (RPN) and Boosted Forest (BF) framework for multispectral input data. Xu \etal\cite{xu2017learning} design a cross-modal representation learning framework to overcome adverse illumination conditions. 

However, most existing methods are employed under the full alignment assumption, hence directly fuse features of different modalities in the corresponding pixel position. This not only hinders the usage of the weakly aligned dataset (\eg CVC-14 \cite{gonzalez2016pedestrian}), but also restricts the further development of multispectral pedestrian detection, which is worthy of attention but still exhibits a lack of study. 

\textbf{Weakly Aligned Image Pair} Weakly aligned image pair is a common phenomenon in multispectral data, since images from different modalities are usually collected and processed independently. A common paradigm to address this problem is to conduct image registration (\ie spatial alignment) \cite{zitova2003image, brown1992survey, dawn2010remote, maintz1998survey} as preprocessing. It geometrically aligns two images: the \textit{reference} and \textit{sensed} images, which can be considered as an image-level solution for the position shift problem. The standard registration includes four typical processes: feature detection, mapping function design, feature matching, and image transformation and re-sampling. Though well-established, the image registration mainly focuses on the low-level transformation of the whole image, which actually introduces time-consuming preprocessing and disenables the CNN-based detector to be trained in an end-to-end way.

\section{Motivation}
\label{SEC3} 
To provide insights into the position shift problem in weakly aligned image pairs, we start with our analysis of the KAIST \cite{hwang2015multispectral} and CVC-14 \cite{gonzalez2016pedestrian} multispectral pedestrian dataset. Then we experimentally study how the position shift problem impacts the detection performance. 
 
\subsection{Important Observations} 
From the multispectral image pairs and the corresponding annotations in the KAIST and CVC-14 dataset, several issues can be observed.

\begin{figure}
\subfigure[]{
\begin{minipage}[t]{0.53\linewidth}
\centering
\includegraphics[width=1.74in]{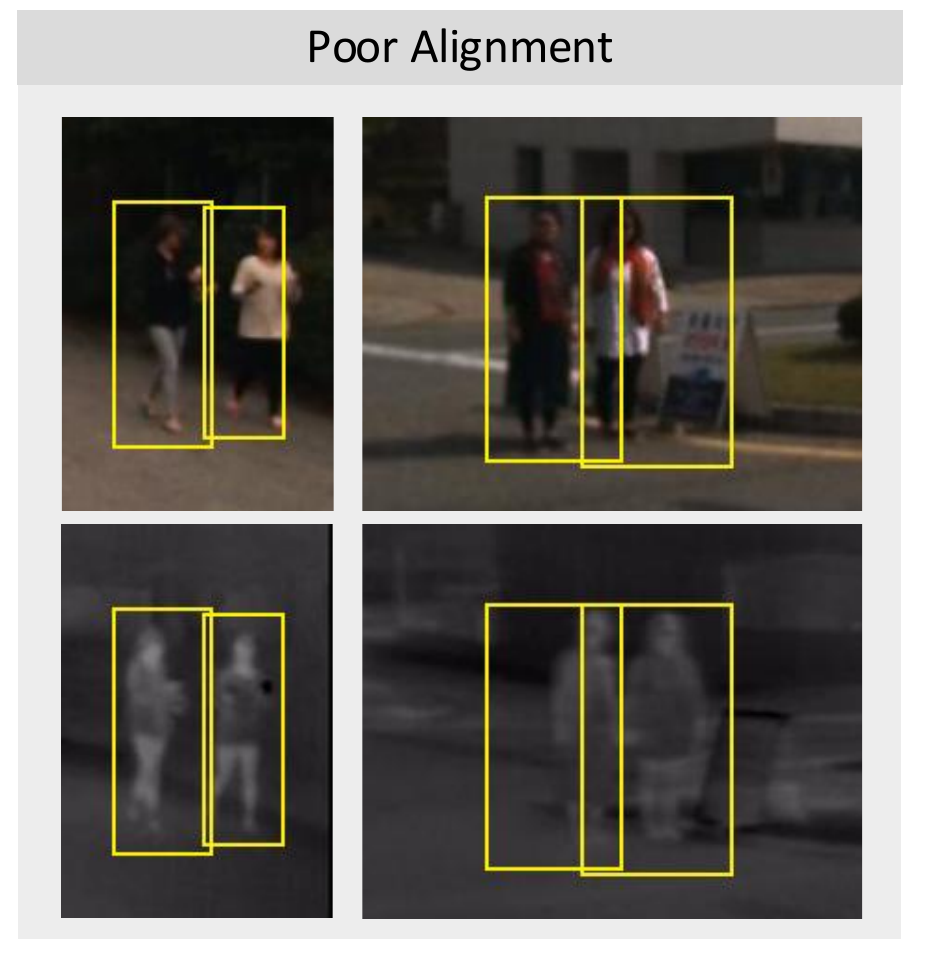}  
\end{minipage}%
\label{fig:side:a}}%
\subfigure[]{
\begin{minipage}[t]{0.45\linewidth}
\centering
\includegraphics[width=1.41in]{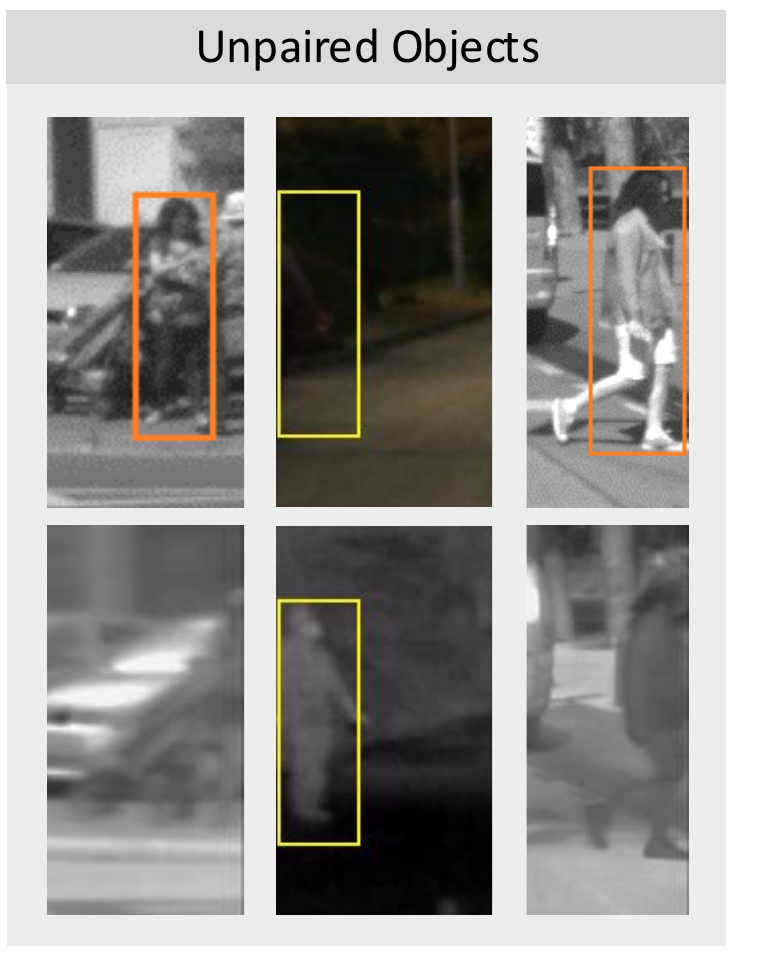}
\end{minipage}%
\label{fig:side:b}}
\caption{The visualization examples of ground truth annotations in the KAIST (boxes in yellow) and CVC-14 (boxes in orange) dataset. Image patches are cropped on the same position of color-thermal image pairs.}%
\label{figure-data-instance}                     
\end{figure}

\textbf{Weakly Aligned Features} As illustrated in Figure \ref{fig:side:a}, the weakly aligned color-thermal image pairs suffer from position shift problem, which makes it unreasonable to directly fuse the feature maps of different modalities.

\textbf{Localization Bias} 
Due to the position shift problem, the annotation must be refined to match both modalities, see Figure \ref{fig:side:a}.
One way is to adopt larger bounding boxes, encompassing pedestrians of both modalities, but generating too big bounding boxes for each modality. Another remedy is to only focus on one particular modality, while introducing bias for another modality.

\textbf{Unpaired Objects} Since the image pair of two modalities may have different field-of-views due to bad camera synchronization and calibration, some pedestrians exist in one modality but are truncated/lost in another, see Figure \ref{fig:side:b}. Specifically, $12.5\%$ ($2,245$ of $18,017$) of bounding boxes are unpaired in the CVC-14 \cite{gonzalez2016pedestrian} dataset.

\subsection{How the Position Shift Impacts?}
\label{sec3.2}
To quantitatively evaluate how the position shift problem influences the detection performance, we conduct experiments on the relatively well-aligned KAIST dataset by manually simulating the position shift.

\textbf{Baseline Detector} We build our baseline detector based on the adapted Faster R-CNN \cite{zhang2017citypersons} framework and adopt the halfway fusion settings \cite{liu2016multispectral} for multispectral pedestrian detection. To mitigate the negative effect of harsh quantization on localization, we use RoIAlign \cite{he2017mask} instead of the standard RoIPool \cite{ren2015faster} for the region feature pooling process. 
Our baseline detector is solid: it has $15.18$ MR on the KAIST reasonable test set, $10.97$ better than the $26.15$ reported in \cite{liu2016multispectral, li2019illumination}. 

\begin{figure}
\subfigure[]{
\begin{minipage}[t]{0.50\linewidth}
\centering
\includegraphics[width=1.56in]{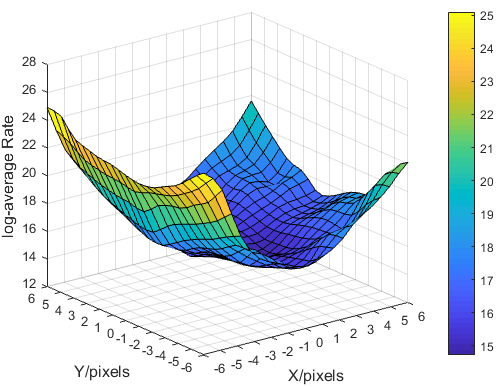}
\label{figure-shifting-1}
\end{minipage}}%
\subfigure[]{
\begin{minipage}[t]{0.50\linewidth}
\centering
\includegraphics[width=1.56in]{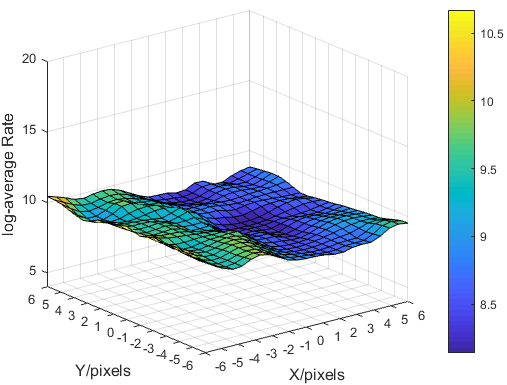}
\label{figure-shifting-2}
\end{minipage}}%
\caption{Surface plot of the detection performances within the position shift experiments. (a) Baseline detector. (b) The proposed approach. Horizontal coordinates indicate different step sizes by which sensed images are shifted along the x-axis and y-axis. Vertical coordinates denote the log-average miss rates (MR) measured on the reasonable test set of KAIST dataset, lower is better.} 
\label{figure-shifting}                     
\end{figure}

\textbf{Robustness to Position Shift} In the testing phase, we fix the thermal image but spatially shift the color image along x-axis and y-axis. The shift pixel is selected in $\{(\Delta x,\Delta y)~|~\Delta x,\Delta y \in [-6,6];\Delta x,\Delta y \in Z \}$, which contains a total of $169$ shift modes. As shown in Figure \ref{figure-shifting-1} and Table \ref{table-num-shifting}, the performance dramatically drops as the absolute shift values increase. Especially, the worst case $(\Delta x,\Delta y)=(-6,6)$ suffers $\sim\textbf{65.3\%}$ relative performance decrement, \ie from $15.18$ MR to $25.10$ MR. Interestingly, when the image is shifted to a specific direction ($\Delta x=1$, $\Delta y=-1$), a better result is achieved ($15.18$ MR to $14.68$ MR), which indicates that we can improve the performance by appropriately handling the position shift problem.

\begin{table}[!hbp]
\begin{center}
\linespread{1.09}\selectfont
\begin{tabular}{|c|c|c|c|c|c|c|}
\hline
\multicolumn{2}{|c|}{ \multirow{2}*{$\Delta$MR ($\%$)} }& \multicolumn{5}{c|}{$\Delta x$} \\
\cline{3-7}
\multicolumn{2}{|c|}{}&-6 &-4 &-1 &0 &1\\
\hline
\multirow{5}*{$\Delta y$}&1 &$\downarrow$~6.55 &$\downarrow$~2.62 &$\downarrow$~0.31 &$\downarrow$~0.14 & $\downarrow$~0.49\\
\cline{2-7}
&0 &$\downarrow$~6.32 &$\downarrow$~2.41 &$\downarrow$~0.21 &\textcolor[rgb]{0,0,1}{0$_{(15.18)}$} & $\downarrow$~0.14\\
\cline{2-7}
&-1 &$\downarrow$~7.19 &$\downarrow$~2.58 &$\downarrow$~0.19 &$\uparrow$~0.25 &$\uparrow$~0.50\\
\cline{2-7}
&-4 &$\downarrow$~8.27 &$\downarrow$~4.01 &$\downarrow$~0.84 &$\uparrow$~0.18 & $\downarrow$~0.22\\
\cline{2-7}
&-6&$\downarrow$~9.92 &$\downarrow$~5.27 &$\downarrow$~1.79 &$\downarrow$~1.37 &$\downarrow$~1.21\\
\hline
\end{tabular}
\end{center}
\caption{Numerical results of the position shift experiments. The scores are corresponding with the results in Figure \ref{figure-shifting-1}. Result in the origin is highlighted in blue, $\downarrow$ refers to performance drop and $\uparrow$ on the contrary.}
\label{table-num-shifting}
\end{table}

\begin{figure}
\subfigure[KAIST]{
\begin{minipage}[t]{0.49\linewidth}
\centering
\includegraphics[width=1.52in]{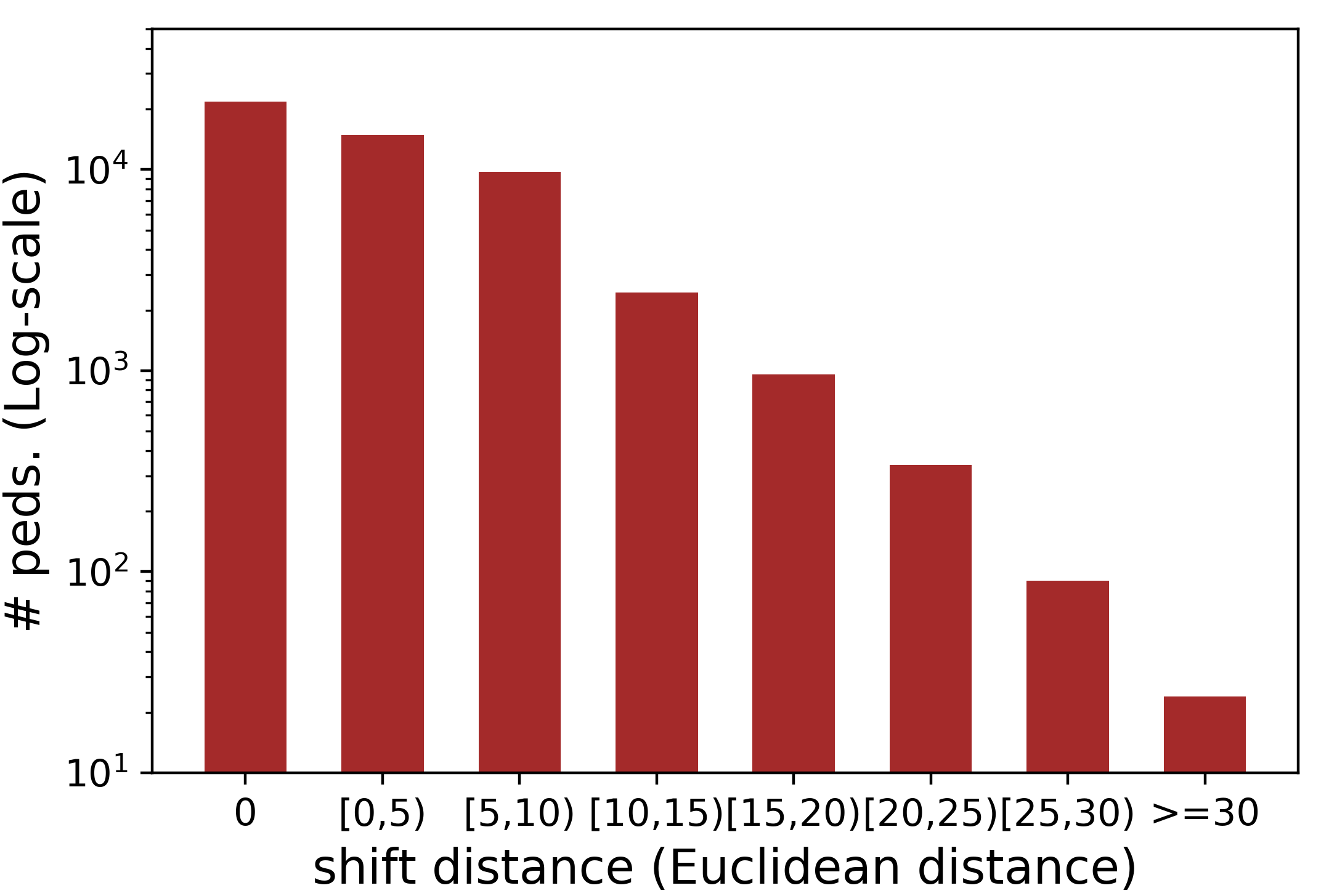}
\label{figure-stat-shifting-1}
\end{minipage}}%
\subfigure[CVC-14]{
\begin{minipage}[t]{0.51\linewidth}
\centering
\includegraphics[width=1.56in]{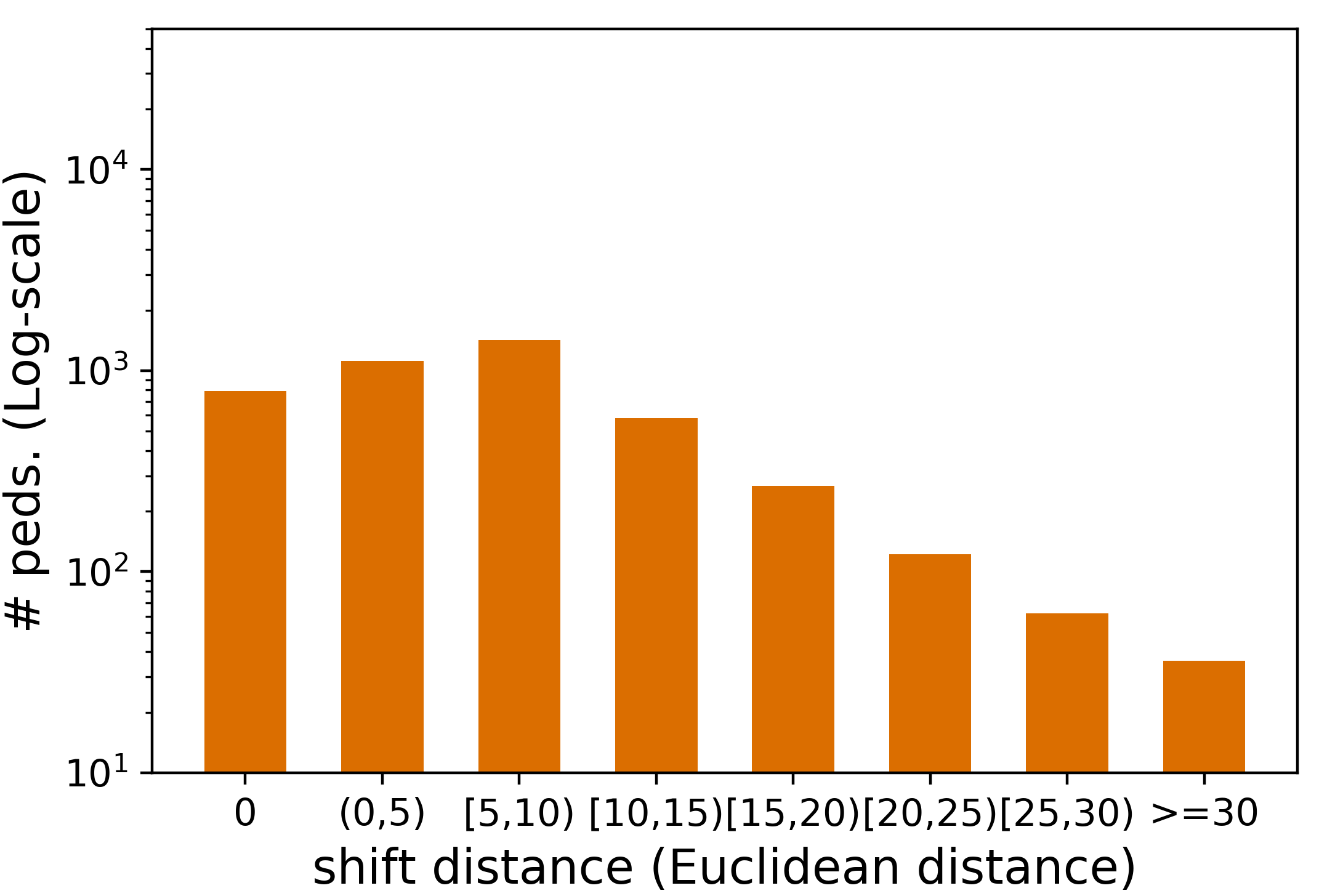}
\label{figure-stat-shifting-2}
\end{minipage}}%
\caption{The statistics of bounding box shift within color-thermal image pairs in KAIST and CVC-14 dataset.} 
\label{figure-stat-shifting}                     
\end{figure}

\begin{figure*}[!t]
\centering
\includegraphics[width=6.25in]{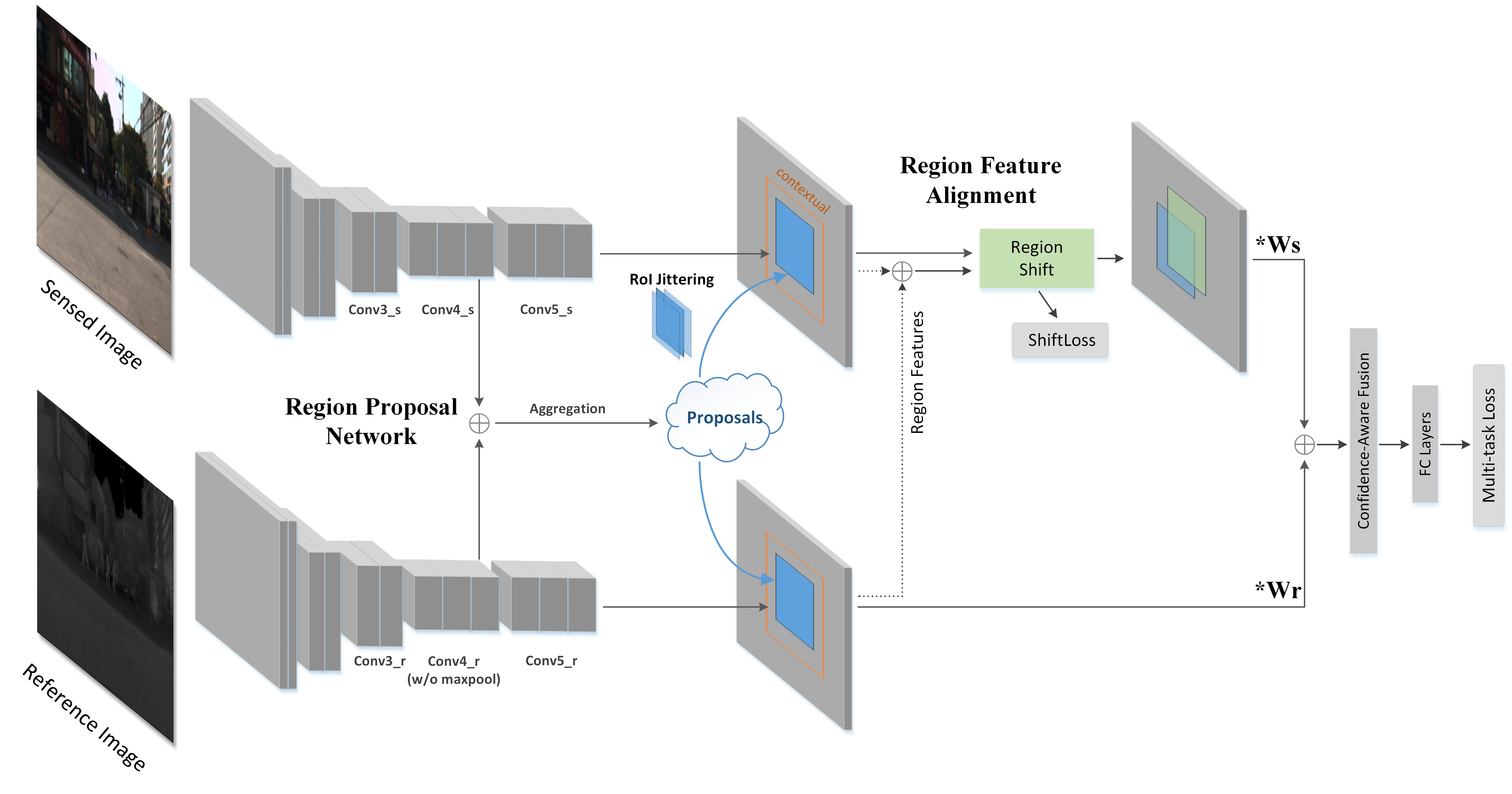}
\caption{The network structure of Aligned Region CNN (AR-CNN). We adopt the two-stream framework to deal with color-thermal inputs. Given a pair of images, numerous proposals are generated and aggregated by the Region Proposal Network, then the Region Feature Alignment module is introduced to align the region features. 
After alignment, the region features of color and thermal feature maps are pooled respectively, then the confidence-aware fusion method is performed.
}
\label{figure-RFA}
\end{figure*}

\section{The Proposed Approach}
\label{sec4}

This section introduces the proposed KAIST-Paired annotation (Section \ref{sec4.1}) and Aligned Region CNN. The architecture of AR-CNN is shown in Figure \ref{figure-RFA}, which consists of the region feature alignment module (Section \ref{sec4.2}), the RoI jitter training strategy (Section \ref{sec4.3}) and the confidence-aware fusion step (Section \ref{sec4.4}).

\subsection{KAIST-Paired Annotation}
\label{sec4.1}
In order to address the position shift problem, we first manually annotate the color-thermal bounding boxes pairs on each modality by the following principles:
\begin{itemize}
\item{Localizing both modalities. The pedestrians are localized in both color and thermal images, aiming to clearly indicate the object locations on each modality.}
\item{Adding relationships. A unique index is assigned to each pedestrian, indicating the pairing information between modalities.}
\item{Labelling the unpaired objects. The pedestrians that only appear in one modality are labelled as ``unpaired'' to identify such situation.}
\item{Extreme case. If the image quality of one modality is beyond human vision, \eg color image under extremely bad illumination, we make the bounding box of color modality consistent with that in the thermal modality.}
\end{itemize}

\textbf{Statistics of KAIST-Paired}
From the new KAIST-Paired annotation, we can get the statistics information of shift distance in the original KAIST dataset. As illustrated in Figure \ref{figure-stat-shifting-1}, more than half of the bounding boxes have the position shift problem, and the shift distance mostly ranges from $0$ to $10$ pixels.

\subsection{Region Feature Alignment}
In this subsection, we propose the Region Feature Alignment (RFA) module to predict the shift between two modalities. Note that the position shift is not simply affine transformation and depends on the cameras. Furthermore, the shift distance varies from pixels to pixels, always small in the center and large in the edge. As a result, the shift prediction and alignment process is performed in a region-wise way.

\label{sec4.2}
\textbf{Reference and Sensed Modality} We introduce the concept of the \textit{reference} and \textit{sensed} \cite{zitova2003image, brown1992survey} image into the multispectral setting. In our implementation, we select the thermal image as the reference modality and the color image as the sensed modality.  
During training, we fix the reference modality and perform the learnable feature-level alignment and RoI jitter process on the sensed one. 

\textbf{Proposals Generation} As illustrated in Figure \ref{figure-RFA}, we utilize the Region Proposal Network (RPN) to generate numerous proposals. We aggregate the proposals from both reference feature map (Conv4\_r) and sensed feature map (Conv4\_s) to keep a high recall rate. 

\begin{figure}[!t]
\centering
\includegraphics[width=3in]{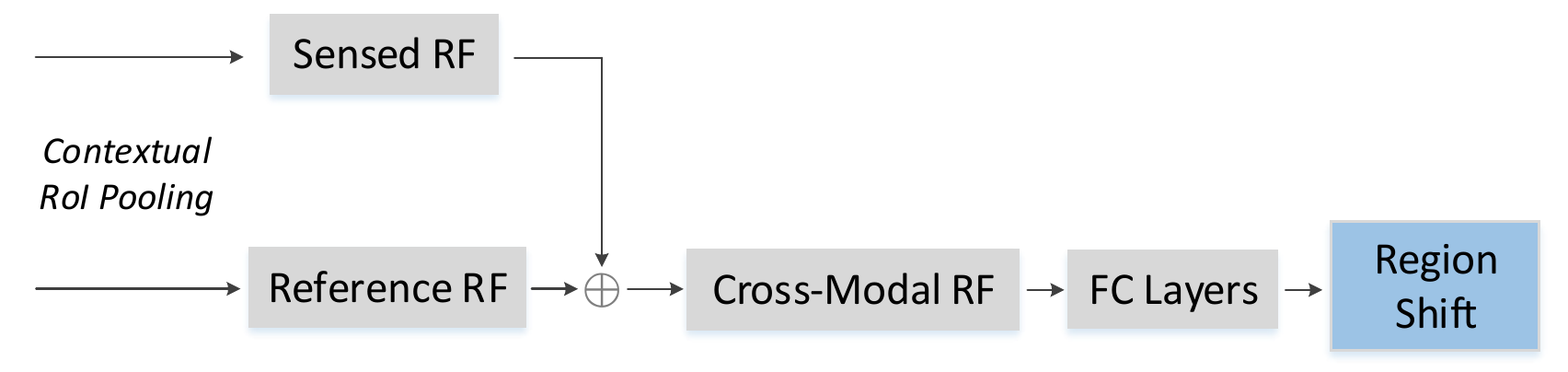}
\caption{Connection scheme of the RFA module. RF denotes region feature and $\oplus$ refers to channel concatenation. The cross-modal region feature is fed into two fully-connected layers to predict this region's shift between two modalities.}
\label{figure-RFA-1}
\end{figure}

\textbf{Alignment Process} The concrete connection scheme of the RFA module is shown in Figure \ref{figure-RFA-1}. Firstly, given several proposals, this module enlarges contextual RoIs to encompass sufficient information of regions. For each modality, the contextual region features are pooled into a small feature map with a fixed spatial extent of $H\times W$ (\eg $7\times 7$). Secondly, the feature map from each modality is then concatenated to get the multimodal representation. From this representation, two consecutive fully connected layers are used to predict the shift targets (\ie $t_{x}$ and $t_{y}$) of this region, so that the new coordinates of the sensed region is predicted. Finally, we re-pool the sensed feature map on the new region to get aligned feature representation with the reference modality. 
Since we have access to the annotated bounding boxes pairs on both modalities, the ground truth shift targets of the two region features can be calculated as follow:

\begin{equation}
\begin{aligned}
&~t^{*}_{x} = (x_{s}-x_{r}) / w_r ~~~~~ t^{*}_{y} = (y_{s}-y_{r}) / h_{r}\\
\end{aligned}
\label{equation-targets}
\end{equation}

In Equation \ref{equation-targets}, $x$, $y$ denote the center coordinates of the box, $w$ and $h$ indicate the width and height of the box . Variables $x_{s}$, $x_{r}$ are for the sensed and reference ground truth box respectively, $t^{*}_{x}$ is the shift target for $x$ coordinate, and likewise for $y$.

\textbf{Multi-task Loss} Similar to Fast R-CNN \cite{girshick2015fast}, we use the smooth L1 loss as the regression loss to measure the accuracy of predicted shift targets, \ie ,

\begin{equation}
\begin{aligned}
L_{shift}( \{p_{i}^{*}\}, \{t_{i}\}, \{t_{i}^{*}\}) = \\ \frac{1}{N_{shift}} \sum _{i=1} ^{n} p_{i}^{*} \mathrm{smoothL_{1}}(t_{i} - t_{i}^{*})
\end{aligned}
\end{equation}
where $i$ is the index of RoI in a mini-batch, $t_{i}$ is the predicted shift target, $p^{*}_{i}$ and $t^{*}_{i}$ are the associated ground truth class label (pedestrian $p^{*}_{i}=1$ \vs background $p^{*}_{i}=0$) and shift target of the i-th sensed RoI. $N_{shift}$ is the total number of ground truth objects to be aligned (\ie $N_{shift}=n$). 

For each training example, we minimize an objective function of Fast R-CNN which is defined as follow:
\begin{equation}
\begin{aligned}
L(\{p_{i}\}, \{t_{i}\}, \{g_{i}\}, \{p_{i}^{*}\}, \{t_{i}^{*}\}, \{g_{i}^{*}\}) = L_{cls}(\{p_{i}\},\{p_{i}^{*}\})\\ + \lambda L_{shift}(\{p_{i}^{*}\}, \{t_{i}\}, \{t_{i}^{*}\}) + L_{reg}(\{p_{i}^{*}\}, \{g_{i}\}, \{g_{i}^{*}\})\\ 
\end{aligned}
\end{equation}
where $p_{i}$ and $g_{i}$ are the predicted confidence and coordinates of the pedestrian, $p_{i}^{*}$ and $g_{i}^{*}$ are the associated ground truth label and the reference ground truth coordinates. Here the two terms $L_{shift}$ and $L_{reg}$ are weighted by a balancing parameter $\lambda$. In our current implementation, we set $\lambda = 1$, and thus the two terms are roughly equally weighted. For the RPN module, the loss function is defined as in the literature \cite{ren2015faster}.

\begin{figure}
\subfigure[Reference image]{
\begin{minipage}[t]{0.45\linewidth}
\centering
\includegraphics[width=1.48in]{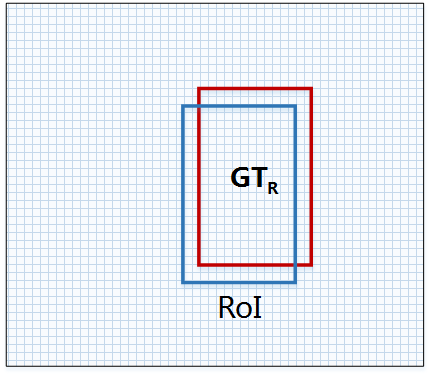}
\label{fig:roij:side:a}
\end{minipage}}%
\subfigure[Sensed image]{
\begin{minipage}[t]{0.45\linewidth}
\centering
\includegraphics[width=1.48in]{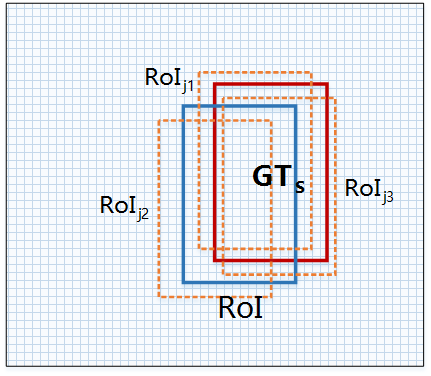}
\label{fig:roij:side:b}
\end{minipage}}%
\caption{Illustration of the RoI jitter strategy. Red boxes denote the ground truths, $\mathrm{GT}_{R}$ and $\mathrm{GT}_{S}$ stand for the reference and sensed modality respectively. Blue boxes represents the RoIs, \ie the proposals, which are shared by both modalities. $\mathrm{RoI}_{j1}$, $\mathrm{RoI}_{j2}$, and $\mathrm{RoI}_{j3}$ are three feasible proposal instances after jitter.} 
\label{figure-RoIJ}                     
\end{figure}

\subsection{RoI Jitter Strategy}
\label{sec4.3}

In reality, the shift patterns are unexpected due to the changes of devices and system settings. 
To improve the robustness to shift patterns, we propose a novel RoI jitter strategy to augment the shift modes. Specifically, the RoI jitter introduces stochastic disturbances to the sensed RoIs and shifts the targets of RFA accordingly, which enriches the patterns of position shift in the training process, as shown in Figure \ref{figure-RoIJ}.

The jitter targets are randomly generated from a normal distribution,
\begin{equation}
\begin{aligned}
t^{j}_{x},~ t^{j}_{y} \sim N(0,\sigma_{0}^{2};&0,\sigma_{1}^{2};0)\\
\end{aligned}
\end{equation}
where $t^{j}$ denotes jitter targets of x-axis and y-axis, and $\sigma$ is the hyperparameter of the radiation extent of jitter.
After, the $\mathrm{RoI}$ jitters to the $\mathrm{RoI}_{j}$ by using the inverse process of bounding box transformation of Equation \ref{equation-targets}.

\textbf{Mini-batch Sampling}
While training the CNN-based detector, a small set of samples is randomly selected. We consistently define the positive and negative examples with respect to the reference modality, since the RoI jitter process is only performed on the sensed modality. Specifically, the RoI pair is treated as positive if the reference RoI has the IoU overlap with reference ground truth box greater than 0.5, and negative if the IoU is between 0.1 and 0.5.

\begin{figure}[!t]
\centering
\includegraphics[width=3.1in]{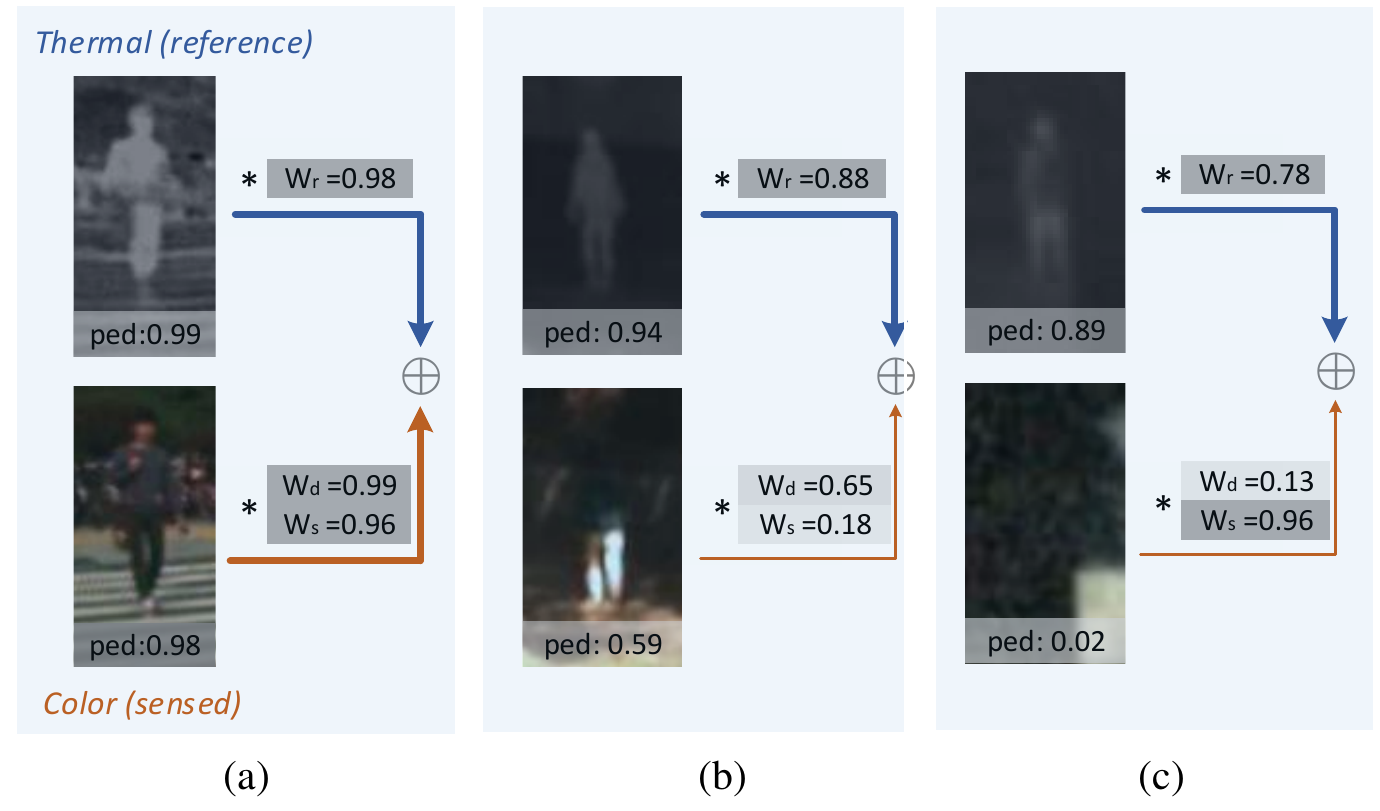}
\caption{Illustration of the confidence-aware fusion method. There are three typical situations: (a) at day time, the color and thermal features are consistent and complementary. (b) under poor illumination, it is difficult to distinguish the pedestrian in color modality, hence we pay more weight on the thermal modality. (c) the pedestrian only exists in the thermal modality due to the position shift, so we depress the color feature.}
\label{figure-CAF}
\end{figure}

\subsection{Confidence-Aware Fusion}
\label{sec4.4}
In around-the-clock operation, modalities provide variational qualities of information: the color data is discriminable at day time but fades at night; the thermal data presents clear human shape throughout the day and night while loses fine visual details (\eg clothing). 
The naive fusion of features from different modalities is not appropriate since we want the detector to pay more attention to reliable modality. 
To this end, we propose a confidence-aware fusion method to make full use of the characteristics between two different modalitites via re-weighting their features, and select the more informative features while suppressing less useful ones. 

As shown in Figure \ref{figure-CAF}, the confidence-aware module has a two-stream architecture and fuses the feature maps from different modalities. 
This module adds a branch for each modality, which is composed of two fully connected layers for the confidence prediction. We calculate two confidence weights: $\mathrm{W}_{r} = |p^{1}_{r}-p^{0}_{r}|$, $\mathrm{W}_{s} = |p^{1}_{s}-p^{0}_{s}|$, in which $p^{1}$ and $p^{0}$ denote the probability of pedestrian and background, $r$ and $f$ refer to the reference and sensed modality, respectively. Then, we use multiplication to perform feature re-weighting (see Figure \ref{figure-CAF}) on the input feature maps to select more reliable features for estimation.

\textbf{Unpaired Objects} During training, since the unpaired objects only exist in one modality, treating them as either background or foreground will lead to ambiguous classification. To mitigate this feature ambiguity, we calculate a disagreement weight, $\mathrm{W}_{d} = 1-|p^{1}_{r} - p^{1}_{s}| = 1-|p^{0}_{r} - p^{0}_{s}|$, and perform re-weighting on the sensed features, \ie the sensed feature will be depressed if it provides a contradictory prediction with the reference modality.

\section{Experiments}

\begin{table*}[!htbp]
\begin{center}
\linespread{1.14}\selectfont
\begin{tabular}{cccc|c|c|c|c|c|c|c|c|c}
\toprule 
\multicolumn{4}{c|}{\multirow{2}*{Method}} &  \multicolumn{3}{c|}{MR} & \multicolumn{3}{c|}{MR$^{C}$} & \multicolumn{3}{c}{MR$^{T}$}\\
\cline{5-13}
&&& &Day &Night &All &Day &Night &All &Day &Night &All\\

\hline

\multicolumn{4}{c|}{ACF+T+THOG (optimized) \cite{hwang2015multispectral}}&~29.59~ &~34.98~ &~31.34~ &~29.85~ &~36.77~ &~32.01~ &~30.40~ &~34.81~ &~31.90~\\

\multicolumn{4}{c|}{Halfway Fusion \cite{liu2016multispectral}}&~24.88~ &~26.59~ &~25.75~ &~24.29~ &~26.12~ &~25.10~ &~25.20~ &~24.90~ &~25.51~\\

\multicolumn{4}{c|}{Fusion RPN \cite{konig2017fully}}&~19.55~ &~22.12~ &~20.67~ &~19.69~ &~21.83~ &~20.52~ &~21.08~ &~20.88~ &~21.43~\\

\multicolumn{4}{c|}{Fusion RPN+BF \cite{konig2017fully}}&~16.49~ &~15.15~ &~15.91~ &~16.60~ &~15.28~ &~15.98~ &~17.56~ &~14.48~ &~16.52~\\

\multicolumn{4}{c|}{Adapted Halfway Fusion}&~15.36~ &~14.99~ &~15.18~ &~14.56~ &~15.72~ &~15.06~ &~15.48~ &~14.84~ &~15.59~\\

\multicolumn{4}{c|}{IAF-RCNN \cite{li2019illumination}}&~14.55~ &~18.26~ &~15.73~ &~14.95~ &~18.11~ &~15.65~ &~15.22~ &~17.56~ &~16.00~\\

\multicolumn{4}{c|}{IATDNN+IAMSS \cite{guan2019fusion}}&~14.67~ &~15.72~  &~14.95~ &~14.82~ &~15.87~ &~15.14~ &~15.02~ &~15.20~ &~15.08~\\

\multicolumn{4}{c|}{CIAN \cite{zhang2019cross}}&~14.77~ &~11.13~ &~14.12~ &~15.13~ &~12.43~ &~14.64~ &~16.21~ &~9.88~ &~14.68~\\

\multicolumn{4}{c|}{MSDS-RCNN \cite{li_2018_BMVC} }   &~10.60~ &~13.73~ &~11.63~ &~9.91~ &~14.21~ &~11.28~ &~12.02~ &~13.01~ &~12.51~\\

\hline
\multicolumn{4}{c|}{\multirow{1}*{AR-CNN (Ours)}} &~\textbf{9.94}~ &~\textbf{8.38}~ &  ~\textbf{9.34}~ &~\textbf{9.55}~ &~\textbf{10.21}~ &~\textbf{9.86}~ &~\textbf{9.08}~ &~\textbf{7.04}~ &~\textbf{8.26}~\\  
\bottomrule
\end{tabular}
\end{center}
\caption{Comparisons with the state-of-the-art methods on the KAIST dataset. Besides the MR protocol, we also evaluate the detectors on MR$^{C}$ and MR$^{T}$ in the KAIST-Paired annotation.}
\label{table-KAIST-measure}
\end{table*}

In this section, we conduct several experiments on the KAIST \cite{hwang2015multispectral} and CVC-14 \cite{gonzalez2016pedestrian} dataset. We set the more reliable thermal input as the reference image and color input as the sensed one, the opposite configuration is discussed in the supplementary material. All methods are evaluated on the ``reasonable'' setup \cite{dollar2012pedestrian}.

\subsection{Dataset}
\textbf{KAIST} The popular KAIST dataset \cite{hwang2015multispectral} contains $95,328$ color and thermal image pairs with $103,128$ dense annotations and $1,182$ unique pedestrians. It is recorded in various scenes at day and night to cover the changes in light conditions.  Detection performance is evaluated on the test set, which consists of $2,252$ frames sampled every $20$th frame from videos.

\begin{table}[h]
\begin{center}
\linespread{1.14}\selectfont
\begin{tabular}{c|c|c|c|c}
\toprule
\multirow{2}*{~}&\multirow{2}*{Method} & \multicolumn{3}{c}{MR} \\
\cline{3-5}
&&Day&Night&All\\

\hline
\multirow{5}*{\rotatebox{90}{Visible}}&\multirow{1}*{SVM \cite{gonzalez2016pedestrian}}&~37.6~&~76.9~&~-~\\

&\multirow{1}*{DPM \cite{gonzalez2016pedestrian}}&~25.2~&~76.4~&~-~\\

&\multirow{1}*{Random Forest \cite{gonzalez2016pedestrian}}&~26.6~&~81.2&~-~\\

&\multirow{1}*{ACF \cite{Park2018Unified}}&~65.0~&~83.2&~71.3~\\

&\multirow{1}*{Faster R-CNN \cite{Park2018Unified}}&~43.2~&~71.4~&~51.9~\\

\hline

\multirow{5}*{\rotatebox{90}{Visible+Thermal~}}&\multirow{1}*{MACF \cite{Park2018Unified}}&~61.3~&~48.2~&~60.1~\\

&\multirow{1}*{Choi \etal} \cite{choi2016multi} &~49.3~&~43.8~&~47.3~\\

&\multirow{1}*{Halfway Fusion \cite{Park2018Unified}}&~38.1~&~34.4~&~37.0~\\

&\multirow{1}*{Park \etal \cite{Park2018Unified}} &~31.8~&~30.8~&~31.4~\\

\cline{2-5}

&\multirow{1}*{AR-CNN (Ours)}&~\textbf{24.7}~&~\textbf{18.1}~&~\textbf{22.1}~\\
\bottomrule
\end{tabular}
\end{center}
\caption{Pedestrian detection results on the CVC-14 dataset. MR is used to compare the performance of detectors. The first column refers to input modalities of the approach. We use the reimplementation of ACF, Faster R-CNN, MACF, and Halfway Fusion in literature \cite{Park2018Unified}.}
\label{table-cvc}
\end{table}

\begin{table*}[!htbp]
\begin{center}
\linespread{1.14}\selectfont
\begin{tabular}{cccc|c|c|c|c|c|c|c|c|c}
\toprule 
\multicolumn{4}{c|}{\multirow{2}*{Method}} &  \multicolumn{3}{c|}{$S^{0^{\circ}}$} & \multicolumn{2}{c|}{$S^{45^{\circ}}$} & \multicolumn{2}{c|}{$S^{90^{\circ}}$} & \multicolumn{2}{c}{$S^{135^{\circ}}$}\\
\cline{5-13}
&&&&$O$ &$\mu$ &$\sigma$ &$\mu$ &$\sigma$  &$\mu$ &$\sigma$ &$\mu$ &$\sigma$\\

\hline

\multicolumn{4}{c|}{Halfway Fusion \cite{liu2016multispectral}}&~25.51~ &~33.73~ &~7.17~ &~36.25~ &~9.66~ &~28.30~ &~2.26~ &~36.71~ &~9.87~\\

\multicolumn{4}{c|}{Fusion RPN \cite{konig2017fully}}&~21.43~ &~30.12~ &~7.13~ &~31.69~ &~10.60~ &~24.48~ &~1.97~ &~34.02~ &~10.64~\\

\multicolumn{4}{c|}{Adapted Halfway Fusion}&~15.59~ &~ 23.44 ~ &~7.49~ &~26.91~ &~11.55~ &~17.95~ &~2.03~ &~27.26~ &~11.18~\\

\multicolumn{4}{c|}{CIAN \cite{zhang2019cross}}&~14.68~ &~23.64~ &~7.69~ &~24.07~ &~11.50~ &~15.07~ &~1.35~ &~23.98~ &~11.57~\\

\multicolumn{4}{c|}{MSDS-RCNN \cite{li_2018_BMVC}}&~12.51~ &20.96 &~7.87~ &~24.43~ &~11.74~ &~14.42~ &~1.34~ &~24.23~ &~10.99~\\
\hline
\hline
\multirow{5}*{AR-CNN} &  \multicolumn{1}{c}{RFA} & \multicolumn{1}{c}{RoIJ}& \multicolumn{1}{c|}{CAF} & \multicolumn{1}{c|}{~} & \multicolumn{1}{c|}{~} &\multicolumn{1}{c|}{~} & \multicolumn{1}{c|}{~} &\multicolumn{1}{c|}{~} &\multicolumn{1}{c|}{~} & \multicolumn{1}{c|}{~} &\multicolumn{1}{c|}{~} &\multicolumn{1}{c}{~}\\
\cline{2-4}
&&&&~12.94~ &~21.05~ &~7.10~ &~15.80~ &~9.77~ &~13.46~ &~1.04~ &~16.18~ &~6.91~\\
&$\checkmark$ &~ &~&~10.90~ &~11.91~ &~2.81~ &~12.38~ &~2.59~ &~11.00~ &~0.21~ &~12.34~ &~2.27~\\
&$\checkmark$&$\checkmark$&~ &~9.87~ &~11.17~ &~1.20~ &~11.84~ &~1.71~ &~10.27~ &~0.17~ &~11.50~ &~1.34~\\
&$\checkmark$&$\checkmark$ &$\checkmark$ &~\textbf{8.26}~ &~\textbf{9.34}~ &~\textbf{0.95}~ &~\textbf{9.73}~ &~\textbf{1.24}~ &~\textbf{8.91}~ &~\textbf{0.43}~ &~\textbf{9.79}~ &~\textbf{1.04}~\\  

\bottomrule
\end{tabular}
\end{center}
\vspace{-0.5em}
\caption{Quantitative results of the robustness of detectors to position shift on the KAIST dataset. 
$O$ denotes the MR$^{T}$ score at the origin, $\mu$, $\sigma$ represents the mean and standard deviation of MR$^{T}$ scores respectively. In this testing, we reimplement the ACF+T+THOG, Halfway Fusion and Fusion RPN, and use the model provided in \cite{zhang2019cross} and \cite{li_2018_BMVC} for CIAN and MSDS-RCNN.}
\label{table-measure}
\end{table*}

\textbf{CVC-14} 
The CVC-14 dataset \cite{gonzalez2016pedestrian} contains visible (grayscale) plus thermal video sequences, captured by a car traversing the streets at $10$ FPS during day and night time. The training and testing set contains $7,085$ and $1,433$ frames, respectively. Note that even with  post-processing, the cameras are still not well calibrated. As a result, annotations are individually provided in each modality. It is worth noting that the CVC-14 dataset has a more serious position shift problem, see Figure \ref{figure-stat-shifting-2}, which makes it difficult for state-of-the-art methods to use the dataset \cite{xu2017learning, li2019illumination, guan2019fusion, zhang2019cross, cao2019box}.

\subsection{Implementation Details}
Our AR-CNN detector uses VGG-16 \cite{simonyan2014very} as the backbone network, which is pre-trained on the ILSVRC CLS-LOC dataset \cite{krizhevsky2012imagenet}. 
We set the $\sigma_{0}$ and $\sigma_{1}$ of RoI jitter to $0.05$ by default, which can be adjusted to handle wider or narrower misalignment. All the images are horizontally flipped for data augmentation.
We train the detector for 2 epochs with the learning rate of $0.005$ and decay it by $0.1$ for another 1 epoch. 
The network is optimized using the Stochastic Gradient Descent (SGD) algorithm with 0.9 momentum and 0.0005 weight decay.
Multi-scale training and testing are not applied to ensure fair comparisons with other methods. 

As for evaluation, the log miss rate averaged over the false positives per image (FPPI) range of $[10^{-2},10^{0}]$ (MR) is calculated to measure the detection performance, the lower score indicates better performance. Since there are some problematic annotations in the original test set of the KAIST benchmark, we use the widely adopted improved test set annotations provided by Liu \etal \cite{liu2016multispectral}. Besides, based on our annotated KAIST-Paired, we propose the MR$^{C}$ and MR$^{T}$ which indicate the log-average miss rate on color and thermal modality.

\subsection{Comparison Experiments}
\textbf{KAIST.} We evaluate our approach and conduct comparisons with other published methods, as illustrated in Table \ref{table-KAIST-measure}. Our proposed approach achieves $9.94$ MR, $8.38$ MR, and $9.34$ MR on the reasonable day, night, and all-day subset respectively, better than other available competitors (\ie \cite{hwang2015multispectral, liu2016multispectral, konig2017fully, li2019illumination, guan2019fusion, zhang2019cross}). Besides, in consideration of the position shift problem, we also evaluate the state-of-the-art methods with the KAIST-Paired annotation, \ie log-average miss rate associated with color modality (MR$^{C}$) and thermal modality (MR$^{T}$). From Table \ref{table-KAIST-measure} we can see that our AR-CNN detector has greater advantages, \ie $9.86$ \vs $11.28$ MR$^{C}$ and $8.26$ \vs $12.51$ MR$^{T}$, demonstrating the superiority of the proposed approach.

\textbf{CVC-14.} We follow the protocol in \cite{Park2018Unified} to conduct the experiments. Table \ref{table-cvc} shows that the AR-CNN outperforms all state-of-the-art methods, especially on the night subset ($18.1$ \vs $30.8$ MR). This validates the contribution of thermal modality, and demonstrates the performance can be significantly boosted by correctly utilizing the weakly aligned modalities.

\subsection{Robustness to Position Shift}
Following the settings in Section \ref{sec3.2}, we test the robustness of AR-CNN to position shift by evaluating the MR$^{T}$ on KAIST dataset.
Figure \ref{figure-shifting-2} depicts the visual results by a surface plot. 
Compared to the baseline results in Figure \ref{figure-shifting-1}, it can be observed that the robustness to position shift is significantly enhanced, with the overall performance improved. To further evaluate the robustness, we design four metrics\footnote {$\Delta x$ and $\Delta y$ denote the shift pixels, which are selected in the following sets:
\begin{spacing}{0.3}
\begin{eqnarray*}
\begin{aligned}
& S^{0^{\circ}}:\{(\Delta x,\Delta y)~|~\Delta y=0,\Delta x \in [-10,10]; \Delta x \in Z \} \vspace{0ex}\\
& S^{45^{\circ}}:\{(\Delta x,\Delta y)~|~\Delta x= \Delta y ,\Delta x \in [-10,10]; \Delta x \in Z \} \vspace{0ex}\\
& S^{90^{\circ}}:\{(\Delta x,\Delta y)~|~\Delta x= 0 ,\Delta y \in [-10,10]; \Delta y \in Z \} \vspace{0ex}\\
& S^{135^{\circ}}:\{(\Delta x,\Delta y)~|~\Delta x= -\Delta y ,\Delta y \in [-10,10]; \Delta y \in Z \} \vspace{0ex}
\end{aligned}
\end{eqnarray*}
\end{spacing} 
}: $S^{0^{\circ}}$, $S^{45^{\circ}}$, $S^{90^{\circ}}$, $S^{135^{\circ}}$, where the $0^{\circ}$-$135^{\circ}$ indicate the shift directions. 
For each shift direction, we have 21 shift modes, which range from $-10$ to $10$ pixels. 
The mean and standard deviation of those 21 results are calculated and shown in  Table \ref{table-measure}.
It can be observed that our AR-CNN detector achieves the best mean performance and the smallest standard deviation on all metrics, demonstrating the robustness of the proposed approach under diverse position shift conditions.

\subsection{Ablation Study}
In this section, we perform ablation experiments on the KAIST dataset for a detailed analysis of our AR-CNN detector. All the ablated detectors are trained using the same setting of parameters.

\textbf{Region Feature Alignment Module.}
To demonstrate the contribution of the RFA module, we evaluate the performance with and without RFA in Table \ref{table-measure}. We find the RFA remarkably reduces the MR$^{T}$ and the standard deviation under diverse position shift conditions. Specifically, for $S^{45^{\circ}}$, the standard deviation is reduced by a significant $8.53$ (from $9.77$ to $1.24$), and consistent reduction is also observed on the other three metrics.

\textbf{RoI Jitter Strategy.} Based on the RFA module, we further add the RoI jitter strategy and evaluate its contribution. As shown in Table \ref{table-measure}, the RoI jitter further reduces the mean and standard deviation of results and achieves $9.87$ MR$^{T}$ at the origin. Besides, we can see that RoI jitter works more on standard deviation than the performance, which demonstrates that it improves the robustness to shift patterns.

\textbf{Confidence-Aware Fusion Method.}
To validate the effectiveness of the confidence-aware fusion method, we compared performance with and without it. As shown in Table \ref{table-measure}, the newly added confidence-aware fusion method slightly depresses the standard deviation, and further reduces MR$^{T}$ at the origin by $1.61$. This demonstrates that the detection performance can be further improved by confidence-aware fusion, since it helps the network to select more reliable features for adaptive fusion.

\section{Conclusion}
In this paper, a novel Aligned Region CNN method is proposed to alleviate the negative effects of position shift problem in weakly aligned image pairs. Specifically, we design a new region feature alignment module, which predicts the position shift and aligns the region features between modalities. Besides, an RoI jitter training strategy is adopted to further improve the robustness to random shift patterns. Meanwhile, we present a novel confidence-aware fusion method to enhance the representation ability of fused feature via adaptively re-weighting the features. 
To realize our method, we relabel the large-scale KAIST dataset by locating the bounding boxes in both modalities and building their relationships. 
Our model is trained in an end-to-end fashion and achieves state-of-the-art accuracy on the challenging KAIST and CVC-14 dataset. Furthermore, the detector robustness to position shift is improved with a large margin. 
It is also worth noting that our method is a generic solution for multispectral object detection rather than only the pedestrian problem. 
In the future, we plan to explore the generalization of the AR-CNN detector and extend it to other tasks, considering this weakly aligned characteristic is widespread and hard to completely avoid when multimodal inputs are required.

\noindent \textbf{Acknowledgments}. This work was supported by the National Key Research and Development Plan of China under Grant 2017YFB1300202, the NSFC under Grants U1613213, 61627808, 61876178, and 61806196, the Strategic Priority Research Program of Chinese Academy of Science under Grant XDB32050100.

{\small
\bibliographystyle{ieee_fullname}
\bibliography{mybibfile}
}

\end{document}